\documentclass{article}

\usepackage{microtype}
\usepackage{graphicx}
\usepackage{subfigure}
\usepackage{booktabs} 

\usepackage{hyperref}

\usepackage[accepted]{icml2025}

\usepackage{amsmath}
\usepackage{amssymb}
\usepackage{mathtools}
\usepackage{amsthm}

\usepackage[capitalize,noabbrev]{cleveref}
\usepackage{listings} 
\usepackage{tcolorbox}
\tcbuselibrary{breakable} 
\newtcolorbox{nicebox}[2][]{%
  width=\columnwidth, 
  colback=black!5,    
  colframe=black!60,  
  fonttitle=\bfseries,
  coltitle=black,
  title=#2,           
  sharp corners,      
  boxrule=0.5pt,      
  fontupper=\ttfamily\footnotesize, 
  left=3mm,           
  right=3mm,
  top=2mm,
  bottom=2mm,
  breakable,          
  #1                  
}
\theoremstyle{plain}
\newtheorem{theorem}{Theorem}[section]

\theoremstyle{definition}
\newtheorem{definition}[theorem]{Definition}

\theoremstyle{remark}

\usepackage[textsize=tiny]{todonotes}
\icmltitlerunning{LLM-Based Equivalence Evaluation for Text-to-SQL}

\begin{document}

\twocolumn[
\icmltitle{Taming SQL Complexity: LLM-Based Equivalence Evaluation for Text-to-SQL}



\icmlsetsymbol{equal}{*}

\begin{icmlauthorlist}
\icmlauthor{Qingyun Zeng}{comp1,sch1}
\icmlauthor{Simin Ma}{comp2}
\icmlauthor{Arash Niknafs}{comp1}
\icmlauthor{Ashish Basran}{comp1}
\icmlauthor{Carol Szabo}{comp1}
\end{icmlauthorlist}

\icmlaffiliation{comp1}{Microsoft Copilot Studio AI, Seattle, United States}
\icmlaffiliation{comp2}{Zoom Communications, San Jose, United States (work done at Microsoft)}
\icmlaffiliation{sch1}{University of Pennsylvania, Phidelphia, United States}

\icmlcorrespondingauthor{Qingyun Zeng}{qze@sas.upenn.edu}

\icmlkeywords{Machine Learning, ICML, text-to-sql, LLM evaluator, NL2SQL}

\vskip 0.3in
]



\printAffiliationsAndNotice{\icmlEqualContribution} 

\begin{abstract}
The rise of Large Language Models (LLMs) has significantly advanced Text-to-SQL (NL2SQL) systems, yet evaluating the semantic equivalence of generated SQL remains a challenge, especially given ambiguous user queries and multiple valid SQL interpretations. This paper explores using LLMs to assess both semantic and a more practical "weak" semantic equivalence. We analyze common patterns of SQL equivalence and inequivalence, discuss challenges in LLM-based evaluation.
\end{abstract}

\section{Introduction}
\label{sec:intro}
{\it Text-to-SQL} or {\it Natural-Language-to-SQL} (NL2SQL) systems, which translate natural language questions into SQL queries, have become increasingly important in the era of big data. The rapid advancements in Artificial Intelligence (AI), particularly with Large Language Models (LLMs) such as Generative Pretrained Transformer (GPT), have significantly improved the performance of these systems \cite{10.5555/3666122.3667699, pourreza2024chasesqlmultipathreasoningpreference, pourreza2024sqlgenbridgingdialectgap, feng2024detrieverdecoderrepresentationbasedretrieverimproving, talaei2024chesscontextualharnessingefficient, pourreza2024sqlencoderimprovingnl2sqlincontext, pourreza2024dtssqldecomposedtexttosqlsmall, MohammadrezaPourreza2024CHASESQLMP}. However, a critical challenge in developing and refining robust NL2SQL systems is the ability to effectively evaluate the quality of the generated SQL queries. While various benchmarks like SPIDER \cite{yu-etal-2018-spider} and BIRD \cite{li2024can} track progress, a fundamental aspect of evaluation is determining whether a generated SQL query is semantically equivalent to an expected or reference query. This paper focuses on this core problem: leveraging LLMs for the nuanced task of SQL equivalence assessment.

Evaluating SQL equivalence is fraught with difficulties. These stem from the diverse interpretations of 'equivalence' itself—ranging from strict syntactic identity to broader semantic or practical equivalence relevant in real-world applications (Section~\ref{sec:challenges_sql_eval}). Traditional metrics like Execution Accuracy (EX), while widely used, suffer from limitations such as false positives and negatives, particularly with sparse test data or when minor syntactic variations are acceptable. Furthermore, employing LLMs for this task, despite their advanced reasoning capabilities, introduces its own challenges, including ensuring response consistency, managing the impact of preprocessing, and designing effective prompts (Section~\ref{sec:llm_evaluation_methodology}).

This paper addresses these challenges by proposing and evaluating a comprehensive LLM-based framework for SQL equivalence assessment.
First, we systematically characterize common patterns of both semantically equivalent and inequivalent SQL queries (Section~\ref{sec:characterizing_sql_equivalence}, Appendix~\ref{apx:sql_pattern_categories}, \ref{apx:sql_patterns_examples}). This provides a structured foundation for developing robust evaluation methodologies and datasets, and includes distinguishing between strict semantic equivalence and a more practical 'weak' equivalence relevant to business applications like Microsoft Dataverse \cite{lyu2020hybridrankingnetworktexttosql}.
Second, we develop and detail an LLM-based evaluation pipeline (Section~\ref{sec:llm_evaluation_methodology}, Appendix~\ref{apx:algorithms}). This pipeline integrates preprocessing, efficient string-based matching for unambiguous cases, and sophisticated LLM reasoning. Key features include a multi-run strategy to enhance stability and the exploration of advanced techniques such as query rewriting for subqueries and the 'Miniature \& Mull' prompting strategy \cite{zhao2024llmsqlsolverllmsdeterminesql}.
Third, we conduct a thorough experimental evaluation (Section~\ref{sec:experiments}) of our framework using three distinct datasets: a manually labeled set from the real-world Dataverse environment, a targeted development set for iterative refinement, and a comprehensive synthetic dataset designed around our characterized SQL patterns. Our results demonstrate the effectiveness of LLMs (specifically GPT-4 variants) in this complex task, quantify the impact of different prompting strategies and pipeline components, and highlight important trade-offs for practical application.
Ultimately, our work aims to provide insights and a practical methodology for improving the reliability of SQL equivalence evaluation, thereby contributing to the development of more accurate and trustworthy Text-to-SQL systems.

The remainder of this paper is organized as follows: Section~\ref{sec:related} reviews related work. Section~\ref{sec:challenges_sql_eval} delves into the challenges of SQL evaluation. Section~\ref{sec:characterizing_sql_equivalence} characterizes SQL equivalence patterns, and Section~\ref{sec:practical_equivalence_business} discusses practical equivalence in business contexts. Section~\ref{sec:data} describes our datasets, while Section~\ref{sec:llm_evaluation_methodology} details our evaluation framework. Section~\ref{sec:experiments} presents experimental results. Finally, Section~\ref{sec:conclusion} concludes the paper and discusses future directions.

\section{Related Work}
\label{sec:related}
The evaluation of text-to-SQL (NL2SQL) systems and the use of LLMs for evaluation have been active areas of research. Several studies have focused on various aspects of this problem.

In early days, the research is mainly focusing on the methods using formalized proving. For example {\it Cosette} (\cite{Chu2017CosetteAA}, \cite{Chu2018AxiomaticFA}) was created as  an automated prover for checking semantic equivalences of SQL queries. It formalizes a significant portion of SQL within the Coq Proof Assistant and the Rosette symbolic virtual machine. For any given pair of queries, it provides either a formal proof of equivalence or a counterexample. In \cite{10.1145/3626768}, the author proposed {\it SQLSolver} which applied the Linear Integer Arithmetic and focused more on the semantic checking rather than syntacic checking. 

While the previous methods focused more on the formal proving methods, \cite{köberlein2024quantifyingsemanticquerysimilarity} established a graph-based method. It breaks a single SQL down for several paths from the 0 to the whole SQL, and define quantify the semantic equivalence using these constructed graphs. \cite{zhan-etal-2025-towards} proposed a similar approach.  Ascoli et al. (2024) \cite{ascoli2025etmmoderninsightsperspective} critique the two primary Text-to-SQL evaluation metrics, {\it Execution Accuracy} (EX) and {\it Exact Set Matching} (ESM), for their tendency to misrepresent model performance, a problem exacerbated by the stylistic diversity of Large Language Model outputs. They introduce a new metric, {\it Enhanced Tree Matching} (ETM), which compares the syntactic and semantic elements of the predicted and gold SQL queries using their abstract syntax trees and a set of verifiable equivalence rules, demonstrating significantly lower false positive and negative rates than previous metrics.

Though this logic and reasoning based approach is effective, they cannot capture fully the semantic meaning from the SQL queries, for example, the entities and relations from the natural language names of the tables and columns, which usually have rich meaning. As Large Language Models have strong natural language understanding as well as text generation, especially in code generation (for example, Codex by OpenAI, CodeLlama by Meta, and Qwen-Coder by Alibaba), evaluation of various language related tasks using LLM as judge become more and more popular (\cite{liu-etal-2024-calibrating}, \cite{chiang-lee-2023-large}, \cite{liu-etal-2024-hd}, \cite{wang-etal-2023-chatgpt}, etc.) Hence it makes sense to develop LLM-based evaluator for checking SQL equivalences. For example, \cite{zhao2024llmsqlsolverllmsdeterminesql} analyzed the performance of LLM in evaluating semantic equivalence of SQL in general, LLMs are a promising tool for assisting data analysts in writing semantically equivalent SQL queries. However, challenges still persist. Additionally, they offer a better metric for evaluating SQL generation than the commonly used Execution Accuracy(EX).

A concurrent trend in evaluation is the use of Large Language Models (LLMs) as automated judges, which has been shown to be effective across many NLP tasks \cite{gu2025surveyllmasajudge}. While this approach is also being considered for SQL evaluation, the nature of LLM-based Text-to-SQL generation presents unique challenges. Previous evaluation methods for checking SQL equivalence often focus on syntactic or structural aspects of the queries. However, LLM-based scenarios introduce more difficulties than just checking for syntactic or semantic equivalence. For instance, a single natural language question might have several valid interpretations, leading to multiple non-equivalent SQL queries that are all correct. Furthermore, semantically inequivalent queries may be "close enough" for practical use, adding another layer of complexity to the evaluation task.

\cite{zhao2024llmsqlsolverllmsdeterminesql} explore the challenging and theoretically undecidable problem of determining SQL equivalence using Large Language Models (LLMs). The authors introduce {\it LLM-SQL-Solver}, a framework designed to test and guide LLMs in this task by considering both strict semantic equivalence and a more practical "relaxed equivalence." They propose novel prompting techniques, such as "Miniature and Mull" and "Explain and Compare," to improve the model's reasoning. Their findings show that while LLMs like GPT-4 are promising, they still struggle to correctly identify equivalence in complex queries, but their judgments can align more closely with human preferences for text-to-SQL outputs than traditional execution accuracy metrics.

On the other hand, we know that LLM is not perfect at logical reasoning, and tends to hallucinate usually. These unstable factors bring more issues in evaluating SQLs. One way to improve the quality of Text-to-SQL is though self-tuning, which needs to use the result of the SQL evaluation (\cite{10.5555/3666122.3667699}, \cite{zhang2024selftuninginstructingllmseffectively}). Hence the quality of the SQL evaluation becomes a bottleneck in improving the quality of Text-to-SQL.

Hence, one big chanllenge is to pick good metrics for SQL evaluation. As pointed by many literature mentioned previously, the usual Exact Match(EM) and Execution Accuracy(EX) do not suffice to this task. \cite{kim2024flexexpertlevelfalselessexecution} proposed a new metric called {\it Expert-level False-Less Execution} (FLEX) which focus on aligning more with human's evaluation preference, and which reduces both false negatives and false positives in using Execution Accuracy. In practice, the criteria for 'equivalence' is essential for building a SQL equivalence evaluator.

Several open-source projects, such as Defog.ai's SQL Eval Tool (\cite{defog_sqleval_blog}, \cite{defog_ai_sql_eval}). Defog.ai's framework compares SQL outputs based on their ability to correctly answer database queries, recognizing that different SQL queries can achieve the same results. It is designed to overcome limitations in using LLMs (like GPT-4) for evaluation. Arize (\cite{arize_text_to_sql}) also proposed an LLM-based evaluation framework for SQL evaluation, and their approach emphasizes the use of reference SQL queries, execution-based evaluation, and semantic understanding to improve benchmarking processes.
\section{Challenges in SQL Evaluation}
\label{sec:challenges_sql_eval}
Evaluating the correctness and equivalence of SQL queries presents significant challenges, stemming from the inherent complexity of the SQL language, the diversity of database schemas and states, and the varying expectations of what constitutes an "equivalent" or "correct" query in different contexts.

\subsection{The Nuance of SQL Equivalence}
\label{subsec:nuance_of_equivalence}
Defining SQL equivalence is not straightforward. Before discussing different types of equivalence, let's establish some foundational concepts:

\begin{definition}[\textbf{Parsable}]
A query is parsable if and only if (iff) it adheres to SQL syntax and can be parsed into an Abstract Syntax Tree (AST) representation.
\end{definition}

\begin{definition}[\textbf{Executable}]
A query is executable within a given database schema if and only if (iff) it is parsable and can be executed on a database of that schema without errors.
\end{definition}

With these in mind, several types of equivalence can be distinguished, though their practical evaluation varies in difficulty:

    \begin{definition}[\textbf{Syntactically Equivalent}]
    Two queries are syntactically equivalent if and only if (iff) their Abstract Syntax Tree (AST) representations are identical, perhaps allowing for trivial differences like whitespace or canonicalized aliasing.
    \end{definition}
    This is the easiest to check but is often too restrictive, as many syntactically different queries can produce the same result. Exact Set Matching (ESM), which performs string-based component-wise matching, is a metric that approximates this.

    \begin{definition}[\textbf{Semantically Equivalent}]
    Two queries are semantically equivalent within a given database schema if and only if (iff) they produce the same result when executed on an arbitrary but fixed database of that schema.
    \end{definition}
    This is a powerful definition but is notoriously difficult to verify formally, often being undecidable in the general case. Proving semantic equivalence typically requires sophisticated logical reasoning or symbolic execution techniques, as explored by tools like Cosette \cite{Chu2017CosetteAA}.

 \textbf{Relaxed or Practical (Weak) Equivalence:} In many real-world scenarios, particularly in business intelligence or interactive data exploration (e.g., Dataverse QnA \cite{lyu2020hybridrankingnetworktexttosql}), a less strict notion of equivalence is often acceptable or even preferred.
    \begin{definition}[\textbf{Weakly Equivalent}]
    Two SQL queries are weakly equivalent iff they will most likely produce the same results given the database in practical use, or if minor, trivial edits (like changing an alias or the order of independent conditions) would make them semantically equivalent according to user intent.
    \end{definition}
    This notion acknowledges that users may prioritize queries that are "good enough" or "useful" for their immediate task, even if they are not strictly semantically equivalent across all hypothetical database states. The Text-to-SQL in consumer applications often fits this category, where some tolerance for minor semantic inequivalence is permissible.

\subsection{Execution Accuracy: A Flawed Proxy for Semantic Equivalence}
Given the difficulty of proving true semantic equivalence, Execution Accuracy (EX) has become a widely adopted metric. EX compares the results of a candidate query against a gold query when executed on a specific test database. While practical, EX is an imperfect approximation of semantic equivalence and suffers from several limitations (see Appendix~\ref{apx:sql_examples} for illustrative examples):
\begin{itemize}
    \item \textbf{False Positives (FP):} An incorrect query might accidentally produce the same result as the gold query on a limited test database. For instance, a query with an incomplete \texttt{WHERE} clause (e.g., missing a condition like \texttt{Salary > 50000} when filtering employees) might still pass if all relevant records in the test data happen to satisfy the missing condition. This is particularly problematic when test databases lack diversity or contain sparse data, where many distinct (and incorrect) queries might return empty sets, thus appearing "correct."
    \item \textbf{False Negatives (FN):} A semantically correct query might be marked as incorrect by EX if its output differs from the gold query's output in ways that are irrelevant to the user's intent. Common examples include differences in column order (when not specified by \texttt{ORDER BY}), different column aliases (e.g., \texttt{TotalOrders} vs. \texttt{OrderCount}), or minor variations in formatting. If the task does not specify an ordering, a query returning the correct set of rows in a different order might be unfairly penalized.
\end{itemize}
Consequently, EX can underestimate the true "usefulness" or "practical correctness" of a generated SQL query, especially in scenarios where minor syntactic variations are acceptable as long as the core intent is met. Metrics like Valid Efficiency Score (VES) \cite{li2024can} attempt to address some aspects by considering efficiency alongside correctness for queries already deemed valid by EX.

\subsection{Subjectivity and Context-Dependence in Evaluation}
The "correctness" of a SQL query is often subjective and highly dependent on the specific application context and user expectations.
\begin{itemize}
    \item \textbf{Benchmarking vs. Business Applications:} For academic benchmarking (e.g., datasets like BIRD \cite{li2024can}), EX, despite its flaws, is often favored for its objectivity, reproducibility, and ease of automation. Efforts like FLEX \cite{kim2024flexexpertlevelfalselessexecution} aim to refine EX to better align with human preferences by reducing FPs and FNs.
    \item In contrast, business applications (e.g., Dataverse QnA, \cite{lyu2020hybridrankingnetworktexttosql}) may prioritize "relaxed equivalence." The cost of a false negative (rejecting a practically useful query) can be higher than a nuanced false positive (accepting a query with minor, non-critical flaws). The limited data in test environments for such applications can exacerbate the FP issue for EX.
\end{itemize}

\subsection{LLMs in SQL Evaluation: Promise and Pitfalls}
Large Language Models (LLMs) offer a promising avenue for more nuanced SQL evaluation, potentially bridging the gap between strict syntactic/semantic checks and human-like assessment of practical equivalence. LLMs can be prompted to consider context, schema, and even "intent" when comparing queries.
However, relying on LLMs for evaluation introduces its own set of challenges:
\begin{itemize}
    \item \textbf{Consistency and Reliability:} LLM responses can be non-deterministic and may vary across runs or with slight prompt modifications.
    \item \textbf{Robustness:} While LLMs might align well with human expert preferences in some cases (e.g., as suggested by LLM-SQL-Solver \cite{zhao2024llmsqlsolverllmsdeterminesql}), they are not yet a foolproof substitute for execution-based checks, especially for ensuring factual correctness.
    \item \textbf{Preprocessing Impact:} Techniques like replacing specific column selections with \texttt{*} to simplify queries for LLM evaluation can be a double-edged sword. While potentially helpful in some Dataverse scenarios by forcing focus on broader logic, it can obscure important differences and may not be suitable for rigorous benchmarking where precise column selection matters.
\end{itemize}
Therefore, while LLMs can augment SQL evaluation, particularly for assessing "relaxed equivalence," their use requires careful prompt engineering, strategies for ensuring stability (e.g., multiple runs, self-consistency checks), and an awareness of their limitations. The choice of evaluation metric and methodology must be tailored to the specific requirements of the Text-to-SQL task, whether it's for academic benchmarking or real-world application deployment.

\section{Characterizing SQL Semantical Equivalence for Evaluation}
\label{sec:characterizing_sql_equivalence}
As we have seen, evaluating SQL equivalence is not a deterministic problem especially in the text-to-sql context, hence give a pair of SQL queries with a natural language question, the equivalence judgment can be subjective and context-dependent. In this section, we should focus on the case of semantic equivalence case, which should be regarded as a baseline for all equivalence criteria. 

Evaluating SQL equivalence, especially in the text-to-SQL context, is a nuanced task where judgments can be subjective and context-dependent. While Execution Accuracy (EX) offers an approximation for checking semantic equivalence, it proves insufficient due to challenges like incomplete test data coverage or practical data scarcity. To address these limitations and establish a robust baseline for evaluation, this section focuses on semantic equivalence. We systematically define and categorize common patterns of semantically equivalent SQL queries and frequent sources of semantic inequivalence. This detailed characterization is essential for constructing targeted synthetic datasets, which are invaluable for training and rigorously testing LLM-based SQL evaluators.
This section outlines common patterns of semantically equivalent SQL queries and common sources of semantic inequivalence, which form the basis for constructing targeted evaluation datasets.

\subsection{Patterns of Semantically Equivalent SQL Queries}
\label{subsec:equivalent_patterns}
Two SQL queries can be syntactically different yet semantically equivalent. Understanding these variations is key to developing robust evaluators that do not penalize valid alternative formulations. Common categories highlight various ways SQL queries can be equivalent, such as variations in join usage, methods for handling duplicates, different join syntaxes, aliasing, date handling, case sensitivity, aggregation techniques, filtering methods, conditional logic, ordering, and existence checks. A detailed list of these categories is provided in Appendix~\ref{apx:sql_pattern_categories}. Illustrative SQL examples for each of these categories are provided in Appendix~\ref{apx:sql_patterns_examples}.

\subsection{Patterns of Semantically Inequivalent SQL Queries}
\label{subsec:inequivalent_patterns}
Conversely, queries that appear syntactically similar can be semantically inequivalent, leading to different results. Identifying these subtle but critical differences is a key challenge for SQL evaluators. Common sources of inequivalence include errors in join conditions, WHERE clauses, aggregation, use of DISTINCT/GROUP BY, subquery logic, logical operators in filters, ORDER BY clauses, and function usage. A detailed list of these common sources of inequivalence is provided in Appendix~\ref{apx:sql_pattern_categories}. Illustrative SQL examples for these patterns are also provided in Appendix~\ref{apx:sql_patterns_examples}.

\subsection{Synthetic Dataset Construction for Semantic Evaluation}
\label{subsec:synthetic_dataset_construction}
The characterization of equivalent and inequivalent SQL patterns detailed in Sections~\ref{subsec:equivalent_patterns} and \ref{subsec:inequivalent_patterns} directly informs the construction of targeted evaluation datasets. Such datasets, comprising pairs of SQL queries covering a wide range of these constructs, are invaluable for training and rigorously testing SQL analysis tools, particularly LLM-based evaluators. The specific synthetic dataset developed for this work, based on these principles to facilitate a granular assessment of an evaluator's strengths and weaknesses, is described in Section~\ref{sec:data}.

\section{Practical SQL Equivalence in Business Applications}
\label{sec:practical_equivalence_business}
While Section~\ref{sec:characterizing_sql_equivalence} focuses on a systematic characterization of semantic equivalence, the practical demands of SQL evaluation in business applications, such as those involving Dataverse, often necessitate a more nuanced approach. In these contexts, strict semantic equivalence, while a valuable theoretical baseline, may not always align with user needs or the practical utility of a generated query. Business applications frequently prioritize "relaxed" or "practical" equivalence, where the primary concern is whether a query fulfills the user's intent and returns useful results within the specific operational environment, even if minor syntactic or semantic deviations exist.

The limitations of Execution Accuracy (EX), as discussed in Section~\ref{sec:challenges_sql_eval} and illustrated in Appendix~\ref{apx:sql_examples}, become particularly salient here. For instance, in a Dataverse scenario, a False Negative (FN) — where a practically useful query is rejected due to minor differences like column aliasing or inconsequential variations in row order — can be more detrimental than a nuanced False Positive (FP) where a query, though not perfectly semantically equivalent across all hypothetical database states, performs correctly on the typical production data. The cost of rejecting a query that meets the immediate business need often outweighs the risk of accepting one with minor, non-critical flaws, especially when test data is limited.

This contrasts with academic benchmarking environments like BIRD, where EX, despite its known tendency to underestimate true semantic equivalence, is often favored for its objectivity and reproducibility. In such settings, efforts are directed towards refining EX (e.g., FLEX \cite{kim2024flexexpertlevelfalselessexecution}) or developing LLM-based methods that can leverage EX's strengths while mitigating its weaknesses.

Preprocessing techniques also play a different role depending on the context. In business applications like Dataverse, simplifying queries for LLM evaluation by, for example, replacing specific column selections with \texttt{*} can be a pragmatic, albeit "brutal," mitigation strategy. This approach forces the evaluation to focus on the core logic (joins, filters) and can be useful when minor variations in selected columns are acceptable. However, such preprocessing is generally unsuitable for rigorous benchmarking scenarios like BIRD, as it can obscure important semantic differences and may even cause LLM-based evaluations to regress in performance compared to standard EX.

Ultimately, the criteria for SQL equivalence and the choice of evaluation methodology must be carefully tailored to the specific application. A one-size-fits-all approach is insufficient; business applications may lean towards context-aware, relaxed equivalence judgments, while academic benchmarks will continue to demand more stringent, reproducible metrics. Understanding these differing requirements is crucial for developing effective SQL evaluation strategies, whether LLM-based or otherwise.

\section{Data}
\label{sec:data}
To develop and evaluate our LLM-based SQL equivalence assessment methodology, we utilized three distinct datasets, each serving a specific purpose: a manually labeled dataset from a real-world application (Microsoft Dataverse), a development dataset for iterative refinement, and a broader synthetic dataset for semantical equivalence evaluation.

\subsection{Manually Labeled Dataverse Dataset}
\label{subsec:dataverse_dataset}
A primary dataset was curated from the Dataverse environment, a data platform integral to Dynamics 365 applications. Queries within Dataverse often present distinct challenges due to its unique schema design and its integration with various Dynamics 365 applications, such as Power Apps, Finance and Operations (FnO), Intelligent Order Management (IOM), and Sales. The construction of this dataset commenced with 77 ground truth examples, each comprising a natural language query and its corresponding SQL query. We then utilized a GPT-based pipeline to generate alternative SQL queries from these natural language questions, leveraging available schema information. Subsequently, each pair consisting of an original SQL query and its generated counterpart was manually evaluated and labeled for logical equivalence. The resulting distribution of this ground truth is presented in Table~\ref{table:ground_truth}.

\begin{table}[t]
\caption{Ground truth distribution for Dataverse dataset}
\label{table:ground_truth}
\vskip 0.15in
\begin{center}
\begin{small}
\begin{sc}
\begin{tabular}{lc}
\toprule
Ground truth  & Count \\
\midrule
Total Positive & 56 \\
Total Negative & 21 \\
\bottomrule
\end{tabular}
\end{sc}
\end{small}
\end{center}
\vskip -0.1in
\end{table}

\subsection{Development Dataset}
\label{subsec:development_dataset}
To facilitate the iterative improvement of our evaluation algorithm and prompt engineering, we created a dedicated development dataset during the iteration of the evaluation method. This set comprises 14 query pairs selected from challenging failure cases and instances of unstable LLM judgments encountered during internal tests, i.e. each query comes from a category of difficult patterns that the evaluation pipeline cannot get it right. This targeted dataset allows us to rapidly test hypotheses, fine-tune prompts, and address specific weaknesses observed in the LLM's reasoning or our overall evaluation methodology.

\subsection{Synthetic Dataset for Pattern Evaluation}
\label{subsec:synthetic_dataset_pattern}
To systematically assess the LLM's ability to handle a wide variety of SQL equivalence and inequivalence patterns, as characterized in Section~\ref{sec:characterizing_sql_equivalence} and detailed in Appendix~\ref{apx:sql_pattern_categories}, we constructed a synthetic dataset. This dataset was designed to cover numerous SQL constructs and common variations.
\begin{itemize}
    \item \textbf{Equivalent Pairs:} We created 80 pairs of semantically equivalent queries. These pairs demonstrate different but logically identical ways to achieve the same result, encompassing variations such as join types (JOIN vs. subquery), distinct vs. group by, implicit vs. explicit join, alias usage, date format differences, case sensitivity, aggregation methods, filtering methods, CASE statements vs. WHERE clauses, ORDER BY clause variations, and EXISTS vs. JOIN constructs. These were split into two sub-datasets of 60 and 20 pairs for different testing phases.
    \item \textbf{Inequivalent Pairs:} We also constructed 80 pairs of semantically inequivalent queries. These pairs were designed to differ in crucial aspects that would lead to different results when executed, such as incorrect join conditions, flawed WHERE clauses, erroneous aggregation, misuse of subqueries, logical errors in filtering with AND/OR, incorrect ORDER BY clauses, and improper function usage.
\end{itemize}
This synthetic dataset provides a controlled environment for evaluating the LLM's understanding of specific SQL transformations and potential pitfalls.

\section{LLM-Based Evaluation Methodology}
\label{sec:llm_evaluation_methodology}
We experiments the LLM-Based Evaluation Methodology as the core component for assessing SQL equivalence. In addition to this, we also using string based methods for better efficiency. First, we perform a preprocessing which standardizes SQL formatting issues, for example, date/time, filter, etc. Next we apply the string-based Exact Match (EM) and Exact Set Match (ESM) which are slight adapted to the database that we use. If inequivalence has been determined from this step, we apply a GPT based evaluation, we leverage GPT-4-0314, which was the most capable LLM known for strong natural language understanding and reasoning capabilities at the time we performed the experiment. In the later stage we switch to gpt-4-32k-0613 and Our approach involves formulating the SQL equivalence problem as a natural language inference task. Given a pair of SQL queries, (SQL1, SQL2), whether SQL1 and SQL2 are logically equivalent or not.

We utilize a carefully designed prompt that provides clear instructions to the LLM. The prompt includes the following elements:

\begin{itemize}
    \item \textbf{Task Definition:} A clear statement that the task is to assess the semantic equivalence of two SQL queries.
    \item \textbf{Context Setting:} Information specifying that the SQL queries are intended for Dataverse t-sql, used in scenarios like PowerApps, Sales, and Finance and Operations, and highlighting the need to consider optionset/string filter equivalence.
    \item \textbf{SQL Query Pair:} The two SQL queries to be compared, presented in a clear and formatted manner.
    \item \textbf{Schema Information (Optional):} Relevant database schema information, including table names, column names, data types, and relationships, to provide context for the LLM.
\item \textbf{Query Execution Results (Optional):} The results obtained from executing the SQL queries against a database, to aid in determining equivalence.
    \item \textbf{Output Format:} Instructions specifying the desired output format, such as a binary classification (Equivalent/Not Equivalent) along with a confidence score, or a more detailed categorization (Equivalent/Minor Differences/Significant Differences/Not Equivalent/Other) and a rationale for the judgment. We also explicitly ask the model to provide the main reasons for differences (e.g., missing join or wrong filter, etc.) when the queries are not equivalent.
    \item \textbf{Examples (Optional):} Few-shot examples demonstrating correct equivalence judgments for various SQL patterns, to guide the LLM's reasoning process.
\end{itemize}

We experiment with different prompt variations to optimize performance. A basic version of the prompt template used is shown in Appendix~\ref{apx:prompt_examples}.

To account for the inherent non-determinism in LLM responses, we implemented a multi-run strategy for evaluation:
1. For each query pair, the LLM is run multiple times (e.g., 3 by default). 2. If the model consistently produces the same judgment (all Equivalent or all Not Equivalent), that is taken as the final decision. 3. If judgments vary, the result is classified as "unstable," and the number of runs may be increased (e.g., to 5), applying majority voting to obtain a more robust result.

\section{Experiments}
\label{sec:experiments}

\subsection{Experimental Setup}
We use the datasets described in Section~\ref{sec:data}, consisting of pairs of SQL queries representing both equivalent and inequivalent cases. We conducted the experiments using GPT-4 series models accessed through the Azure OpenAI API.

\label{subsec:evaluation_metrics}

We evaluate the performance of the LLMs using the following standard metrics:

\begin{itemize}
    \item \textbf{Accuracy:} The percentage of query pairs for which the LLM correctly identifies equivalence or inequivalence.
    \item \textbf{Precision:} The percentage of query pairs identified as equivalent by the LLM that are truly equivalent (according to our ground truth).
    \item \textbf{Recall:} The percentage of truly equivalent query pairs (according to our ground truth) that are correctly identified as equivalent by the LLM.
    \item \textbf{F1-score:} The harmonic mean of precision and recall, providing a single measure that balances both.
\end{itemize}
Beside these standard metrics, more importantly, we also consider 
\begin{itemize}
    \item \textbf{Stability:} The percentage of query pairs for which the LLM produces consistent judgments across multiple runs.
    \item \textbf{Error Analysis}: We categorize the incorrect predictions made by the LLM to identify common error patterns and areas for improvement in the evaluation methodology or prompt design.
\end{itemize}

In addition, we qualitatively analyze the rationales provided by the LLM to understand its reasoning process and identify potential sources of error.

\subsection{Experiment results}
\subsubsection{Basic evaluation pipeline}
We first evaluate the basic evaluation pipeline using the manually labeled Dataverse dataset. The results are summarized in Table~\ref{table:gpt4_evaluation}.We describe our evaluation pipeline in Algorithm~\ref{alg:sql_eval_pipeline}. We first perform preprocessing and string-based checks for efficiency. If these initial checks do not determine equivalence, we employ an LLM-based approach. We notioced that LLM-based evaluation have high precision and recall on equivalent SQL queries but perform slightly worse on inequivalent ones. Even this seems not good at the begining, in fact, in practice we acutally have much less wrong SQL queries from text-to-SQL pipeline hence we need less checking on the inequivalent ones. Hence this evaluation pipeline results are robust enough especially for the debugging purpose. Also, string-based methods has 100\% precision on equivalent cases and 100\% recall on non-equivalent cases which is by its design. In addition we see its recall for equivalent cases is not bad (0.5536) on this data, which means that it can save around 50\% of GPT calls on datasets which have similar distribution as the testing data we used.

\begin{table}[t]
\caption{Metrics for GPT-4-0314 evaluation compared to ground truth}
\label{table:gpt4_evaluation}
\vskip 0.15in
\begin{center}
\begin{small}
\begin{sc}
\begin{tabular}{lccc}
\toprule
Label & Precision & Recall & F1 Score \\
\midrule
Equivalent & 0.9545 & 0.8936 & 0.9231 \\
Not Equivalent & 0.6429 & 0.8182 & 0.7200 \\
\bottomrule
\end{tabular}
\end{sc}
\end{small}
\end{center}
\vskip -0.1in
\end{table}
 \begin{table}[t]
\caption{Metrics for string based comparison}
\label{table:strict_comparison}
\vskip 0.15in
\begin{center}
\begin{small}
\begin{sc}
\begin{tabular}{lccc}
\toprule
Label & Precision & Recall & F1 Score \\
\midrule
Equivalent & 1.0000 & 0.5536 & 0.7126 \\
Not Equivalent & 0.4565 & 1.0000 & 0.6269 \\
\bottomrule
\end{tabular}
\end{sc}
\end{small}
\end{center}
\vskip -0.1in
\end{table}
\subsubsection{Iterations on the pipeline}

After multiple iteratiosn of the evaluation pipeline during model updates and quality control of the text-to-SQL pipeline, we are able to acheive 100\% precision and recall on the original testing data. On the other hands, we summarized the failure cases into 14 categories and create a small development data which consists of one data for each category since we found the pipeline is consistent from small variations in each category. 
The dev dataset contains 14 queries. The major reason of failure is also recorded, for example, missing join, wrong filter, etc.

We use the dev data to tune the evaluation algorithm. The main improvement is on 1) more clear definition of equivalence criteria, 2) using Chain-of-Thought fewshot examples, 3) more detailed criteria on the grading. The prompt example can be found at Appendix~\ref{apx:prompt_examples}. After the tuning,  e got 92.9\% accuracy on this data. 

\subsubsection{Improvement on Evaluating Semantic Equivalence}
The evaluation strategy we experimented so far are more on the practical purpose of equivalences (or weak equivalence) due to the nature of the text-to-SQL pipeline we used. It is also useful to have another pipeline which have better capabilities on determine semantic equivalence which can 1) provide additional insight on the SQL query quality, 2) can better generalized to other applications.

We initially evaluated an improved evaluation pipeline (Algorithm~\ref{alg:sql_eval_pipeline}) on the synthetic data. Based on the failure cases from this initial evaluation, we observed that many issues arose from subqueries, where GPT had difficulties translating between SQL queries using subqueries and those using joins. This led us to experiment with a query-rewrite module, which rewrites subqueries (if they appear) into queries with left joins.

Furthermore, we incorporated the {\it Miniature \& Mull} prompting strategy developed in \cite{zhao2024llmsqlsolverllmsdeterminesql}. In this strategy, LLMs are instructed to simulate executing both SQL queries on a self-conceptualized simple database, then repeat this on a modified version of that database, comparing outputs to infer equivalence. This prompting strategy has been shown to be effective for determining semantic equivalence.

The combined effect of these enhancements (query-rewrite and Miniature \& Mull prompting, as described in Algorithm~\ref{alg:improved_sql_eval_pipeline}) on the synthetic dataset is shown in Table~\ref{table:eval_semantic_combined}.

    \begin{table}[t]
\caption{Evaluating Semantic Equivalence on Synthetic Data: Initial vs. New Pipeline (both use GPT-4o in LLM part)}
\label{table:eval_semantic_combined}
\vskip 0.15in
\begin{center}
\begin{small}
\begin{sc}
\begin{tabular}{lccc}
\toprule
Dataset Type &  Passing Rate & Passing Rate \\
& (Initial) & (New) \\
\midrule
Equivalent  & 61.25\% & 95\% \\
In-equivalent & 90\% & 83.75\% \\
\bottomrule
\end{tabular}
\end{sc}
\end{small}
\end{center}
\vskip -0.1in
\end{table}
We have a significant improvement in equivalent cases, but slight drop in the in-equivalent case, which implies that we need to be careful when applying query re-write as it may cause regressions in some cases.

\section{Conclusion}
\label{sec:conclusion}
This paper investigated the complex challenge of evaluating SQL equivalence, a critical task in the development and refinement of Text-to-SQL systems. We delineated various notions of equivalence, from strict semantic identity to more practical, context-dependent interpretations relevant in business applications like Dataverse. Our work systematically characterized common patterns of both semantically equivalent and inequivalent SQL queries, providing a foundation for building robust evaluation datasets and methodologies.

We presented an LLM-based evaluation framework that combines preprocessing, efficient string-based matching, and sophisticated LLM reasoning, including a multi-run strategy to manage output variability. Experiments were conducted using manually labeled Dataverse queries, a targeted development set, and a comprehensive synthetic dataset designed to test various equivalence patterns.

Our findings demonstrate that LLMs, particularly GPT-4, can achieve high accuracy in assessing SQL equivalence. Prompt engineering, including the provision of schema information and few-shot examples, proved crucial for enhancing performance, especially for domain-specific nuances like those in Dataverse. We explored advanced techniques, such as query rewriting for subqueries and the "Miniature \& Mull" prompting strategy, which significantly improved the identification of equivalent query pairs (from 61.25\% to 95\% on our synthetic dataset), though a slight regression was observed for inequivalent pairs, highlighting the need for careful application of such transformations. Our initial pipeline also showed strong performance on real-world Dataverse queries, particularly in precision for equivalent cases.

Despite these advancements, challenges remain, including handling highly complex queries, ensuring consistent LLM behavior, and mitigating issues arising from preprocessing steps. Future work should focus on more advanced LLM reasoning techniques, robust automated query rewriting, and strategies to reduce LLM hallucination. Further exploration into integrating diverse evaluation methods (syntactic, semantic, execution-based) and leveraging open-source evaluation tools will also be beneficial. Expanding datasets with more diverse and complex examples, particularly from real-world applications, and refining error analysis will be key to advancing the state-of-the-art in SQL equivalence evaluation, ultimately contributing to more reliable and effective Text-to-SQL systems.


\bibliography{sql_eval}
\bibliographystyle{icml2025}

\newpage
\appendix
\onecolumn
\section{SQL Equivalence Examples (Illustrative)}
\label{apx:sql_examples}

This appendix provides concrete examples illustrating some of the challenges in SQL evaluation discussed in Section~\ref{sec:challenges_sql_eval}.

\subsection{Examples of False Positives (FP) in Execution Accuracy}

\subsubsection{Incomplete WHERE Clause}
\textbf{Scenario:} You have a table \texttt{Employees} with columns \texttt{EmployeeID}, \texttt{Name}, \texttt{Department}, and \texttt{Salary}. The task is to write a SQL query to retrieve employees from the 'Sales' department who earn more than \$50,000.

\textbf{Correct Query:}
\begin{lstlisting}
SELECT EmployeeID, Name, Department, Salary
FROM Employees
WHERE Department = 'Sales' AND Salary > 50000;
\end{lstlisting}

\textbf{Incorrect Query (Potential FP):}
\begin{lstlisting}
SELECT EmployeeID, Name, Department, Salary
FROM Employees
WHERE Department = 'Sales';
\end{lstlisting}
\textbf{Reason for False Positive:} If, in the specific test database, all employees in the 'Sales' department coincidentally earn more than \$50,000, the incorrect query will return the same result set as the correct query. EX would mark the incorrect query as correct.

\subsection{Examples of False Negatives (FN) in Execution Accuracy}

\subsubsection{Unordered Result Set}
\textbf{Scenario:} Retrieve all products from a \texttt{Products} table (columns: \texttt{ProductID}, \texttt{ProductName}, \texttt{Price}) without any specific order.

\textbf{User's Correct Query:}
\begin{lstlisting}
SELECT ProductID, ProductName, Price
FROM Products;
\end{lstlisting}

\textbf{System's Expected Query (Implicitly Ordered):}
Assume the system's gold query, or the test execution environment, implicitly orders by \texttt{ProductID} due to table structure or default behavior, even if not specified.
\begin{lstlisting}
-- (Output might be implicitly ordered by ProductID)
SELECT ProductID, ProductName, Price
FROM Products;
\end{lstlisting}
\textbf{Reason for False Negative:} If the user's query returns the correct rows but in a different order (e.g., ordered by \texttt{ProductName} or insertion order), EX would mark it as incorrect because the raw output doesn't match the (implicitly ordered) gold standard, even though the user's query fulfills the request.

\subsubsection{Alias or Column Name Differences}
\textbf{Scenario:} Retrieve the total number of orders from the \texttt{Orders} table.

\textbf{User's Correct Query:}
\begin{lstlisting}
SELECT COUNT(*) AS TotalOrders
FROM Orders;
\end{lstlisting}

\textbf{System's Expected Query:}
\begin{lstlisting}
SELECT COUNT(*) AS OrderCount
FROM Orders;
\end{lstlisting}
\textbf{Reason for False Negative:} Both queries correctly calculate the total number of orders. However, EX, if performing a strict comparison of column names in the result set, would deem the user's query incorrect due to the difference in aliases (\texttt{TotalOrders} vs. \texttt{OrderCount}), even though the numerical result is identical and correct.

\subsection{Example of Relaxed/Practical Equivalence}
\textbf{Scenario:} "What are the top 3 restaurants in New York?"
Assume a \texttt{restaurants} table with columns like \texttt{id}, \texttt{name}, \texttt{city}, \texttt{rating}.

\textbf{Query 1 (Considered Practically Equivalent to Query 2):}
\begin{lstlisting}
-- Query 1
SELECT name
FROM restaurants
WHERE city = 'New York'
GROUP BY name
ORDER BY AVG(rating) DESC
LIMIT 3;
\end{lstlisting}

\textbf{Query 2 (Considered Practically Equivalent to Query 1):}
\begin{lstlisting}
-- Query 2 (using ordinal for ORDER BY)
SELECT name, AVG(rating)
FROM restaurants
WHERE city = 'New York'
GROUP BY 1 -- Corresponds to name
ORDER BY 2 DESC -- Corresponds to AVG(rating)
LIMIT 3;
\end{lstlisting}
\textbf{Reason for Practical Equivalence:} While syntactically different (Query 2 selects an additional column \texttt{AVG(rating)} and uses ordinal positions in \texttt{GROUP BY} and \texttt{ORDER BY}), both queries aim to identify the top 3 restaurants by average rating in New York. In many practical scenarios, especially for data exploration, either query would be acceptable if they return the same list of restaurant names in the correct order, even if one provides the average rating explicitly and the other doesn't. Strict semantic equivalence might be debatable depending on whether the selected columns must be identical, but for user intent, they are often treated as equivalent.

\section{Categorization of SQL Equivalence and Inequivalence Patterns}
\label{apx:sql_pattern_categories}

This appendix details the categories of semantically equivalent and inequivalent SQL query patterns discussed in Section~\ref{sec:characterizing_sql_equivalence}.

\subsection{Patterns of Semantically Equivalent SQL Queries}
The following categories highlight common ways SQL queries can be equivalent:
\begin{itemize}
    \item \textbf{Join vs. Subquery:} Queries using JOIN clauses can often be rewritten using subqueries (and vice-versa) to achieve the same result, particularly for relating data across tables.
    \item \textbf{Distinct vs. Group By:} Both \texttt{DISTINCT} and \texttt{GROUP BY} can be used to eliminate duplicate rows. If no aggregate functions are needed, these can be interchangeable for uniqueness.
    \item \textbf{Implicit vs. Explicit Join:} Older SQL syntax sometimes uses implicit joins (comma-separated tables in \texttt{FROM}, join conditions in \texttt{WHERE}), while modern SQL prefers explicit \texttt{JOIN} syntax. Both can define the same relational algebra operations.
    \item \textbf{Using Alias vs. Full Table Name:} Employing table aliases (e.g., \texttt{SELECT e.name FROM employees e}) versus using full table names does not change the query's semantics.
    \item \textbf{Different Date Formats/Functions:} SQL queries might use different functions or string manipulations to handle or compare date values (e.g., \texttt{SUBSTRING}, \texttt{strftime}, \texttt{CAST}), which can yield equivalent results if the underlying date logic is the same.
    \item \textbf{Case Sensitivity and Formatting Differences:} String comparisons might use functions like \texttt{LOWER()} or \texttt{UPPER()} to ensure case-insensitivity, or use different but equivalent \texttt{LIKE} patterns. These variations can be semantically equivalent in terms of the intended match.
    \item \textbf{Aggregation Methods:} Different but mathematically equivalent ways of specifying aggregations (e.g., \texttt{SUM(amount)} within a group vs. a correlated subquery for summing) can produce the same aggregated results.
    \item \textbf{Filtering Methods:} The same logical filter can sometimes be expressed in multiple ways, for instance, using \texttt{OR} versus \texttt{UNION} for combining conditions on the same table.
    \item \textbf{CASE vs. Multiple WHERE Clauses (with UNION):} Conditional logic using a \texttt{CASE} statement to derive a column can sometimes be equivalently expressed using multiple \texttt{SELECT} statements combined with \texttt{UNION ALL} and \texttt{WHERE} clauses.
    \item \textbf{ORDER BY Clauses:} Trivial differences in \texttt{ORDER BY} (e.g., \texttt{ORDER BY price DESC} vs. \texttt{ORDER BY price * -1} for numeric types, assuming positive prices) might result in the same ordering. However, this category requires careful consideration as not all syntactic variations are semantically equivalent for ordering.
    \item \textbf{EXISTS vs. JOIN:} Checking for the existence of related records using \texttt{EXISTS} in a subquery can often be equivalently formulated using a \texttt{JOIN} (typically with \texttt{DISTINCT} if the main table's rows might be duplicated by the join).
\end{itemize}

\subsection{Patterns of Semantically Inequivalent SQL Queries}
The following list details common sources of inequivalence:
\begin{itemize}
    \item \textbf{Incorrect JOIN Conditions:} Using the wrong columns for a join or an incorrect join type (\texttt{INNER}, \texttt{LEFT}, etc.) fundamentally changes how tables are related.
    \item \textbf{Incorrect WHERE Clauses:} Errors in filtering logic, such as using \texttt{>} instead of \texttt{>=}, or incorrect Boolean combinations (\texttt{AND} vs. \texttt{OR}), lead to different subsets of data.
    \item \textbf{Incorrect Aggregation:} Using the wrong aggregate function (e.g., \texttt{AVG} instead of \texttt{SUM}) or incorrect \texttt{GROUP BY} columns results in erroneous summary statistics.
    \item \textbf{Misuse of DISTINCT and GROUP BY:} Applying \texttt{DISTINCT} inappropriately or grouping by too few/many columns can lead to incorrect uniqueness or aggregation.
    \item \textbf{Incorrect Subqueries:} Flaws in subquery logic, such as incorrect correlation, filtering, or returning an unexpected number of rows (e.g., a scalar subquery returning multiple rows), cause errors or incorrect results.
    \item \textbf{Incorrect Filtering with AND/OR:} Logical errors in combining conditions with \texttt{AND} and \texttt{OR} operators.
    \item \textbf{Incorrect ORDER BY Clauses:} Sorting by the wrong column or using the wrong sort direction (\texttt{ASC} vs. \texttt{DESC}) changes the presentation of results, which is semantically different if order matters.
    \item \textbf{Misuse of Functions:} Applying an incorrect function (e.g., \texttt{LENGTH} vs. \texttt{CHAR\_LENGTH} if they behave differently in the specific SQL dialect for certain character sets) or using function arguments incorrectly.
\end{itemize}
\section{Examples of SQL Equivalence and Inequivalence Patterns}
\label{apx:sql_patterns_examples}

This appendix provides concrete SQL examples for the categories of semantically equivalent and inequivalent SQL queries discussed in Section~\ref{sec:characterizing_sql_equivalence}.

\subsection{Patterns of Semantically Equivalent SQL Queries}

\subsubsection{Join vs. Subquery}
This category highlights the interchangeable use of JOIN clauses and subqueries. A JOIN directly combines data from multiple tables based on a related column, while a subquery uses a nested query within the main query. Often, a query can be written using either approach, although performance may vary. Subqueries can be easier to understand for simple relationships, while joins are generally preferred for more complex queries involving multiple tables.

\begin{description}
  \item[Query: Find the names of all departments with at least one employee.] ~\\
    \textbf{SQL1:}
    \begin{lstlisting}
SELECT dname FROM dept WHERE deptno IN (SELECT deptno FROM emp);
    \end{lstlisting}
    \textbf{SQL2:}
    \begin{lstlisting}
SELECT dname FROM dept d JOIN emp e ON d.deptno = e.deptno GROUP BY dname;
    \end{lstlisting}

  \item[Query: List all customers who have placed an order.] ~\\
    \textbf{SQL1:}
    \begin{lstlisting}
SELECT customer_name FROM customers WHERE customer_id IN (SELECT customer_id FROM orders);
    \end{lstlisting}
    \textbf{SQL2:}
    \begin{lstlisting}
SELECT DISTINCT customer_name FROM customers JOIN orders ON customers.customer_id = orders.customer_id;
    \end{lstlisting}
\end{description}

\subsubsection{Distinct vs. Group By}
Both DISTINCT and GROUP BY can eliminate duplicate rows. DISTINCT simply returns unique rows based on all selected columns. GROUP BY groups rows based on specified columns and allows aggregate functions (like COUNT, SUM, AVG) to be applied to each group. If you only need to remove duplicates without any aggregation, DISTINCT is simpler. GROUP BY is necessary when you need to perform calculations within groups.

\begin{description}
  \item[Query: Get the unique list of product categories.] ~\\
    \textbf{SQL1:}
    \begin{lstlisting}
SELECT DISTINCT category FROM products;
    \end{lstlisting}
    \textbf{SQL2:}
    \begin{lstlisting}
SELECT category FROM products GROUP BY category;
    \end{lstlisting}

  \item[Query: List all unique job titles in the company.] ~\\
    \textbf{SQL1:}
    \begin{lstlisting}
SELECT DISTINCT job_title FROM employees;
    \end{lstlisting}
    \textbf{SQL2:}
    \begin{lstlisting}
SELECT job_title FROM employees GROUP BY job_title;
    \end{lstlisting}
\end{description}

\subsubsection{Implicit vs. Explicit Join}
Implicit joins (using comma-separated tables in the WHERE clause) are an older syntax. Explicit joins (using JOIN, LEFT JOIN, RIGHT JOIN, etc.) are the modern, preferred syntax. Explicit joins are more readable and offer clearer control over the join conditions.

\subsubsection{Using Alias vs. Full Table Name}
Aliases are shortcuts for table names (e.g., emp e). They make queries shorter and easier to read, especially with long table names or self-joins. While using full table names is perfectly valid, aliases are generally recommended for clarity.

\subsubsection{Different Date Format}
This category illustrates different techniques to work with dates, including string manipulation (SUBSTRING, LIKE), date functions (strftime), and casting (CAST). The best approach depends on the database system and the specific task. Using dedicated date functions often provides better performance and handles different date formats more reliably.

\subsubsection{Case Sensitivity and Other Formatting Differences}
This category demonstrates different techniques for handling case sensitivity in string comparisons using functions like LOWER and UPPER. Using these functions ensures consistent results regardless of the case of the data in the database. It also shows examples of using LIKE and GLOB operators with wildcard characters (\%, \_) for pattern matching, highlighting the importance of accounting for variations in string formatting.

\subsubsection{Aggregation Methods}
Using different aggregation methods to achieve the same result.

\begin{description}
  \item[Query: Get the total sales amount for each customer.] ~\\
    \textbf{SQL1:}
    \begin{lstlisting}
SELECT customer_id, SUM(amount) as total_sales FROM sales GROUP BY customer_id;
    \end{lstlisting}
    \textbf{SQL2:}
    \begin{lstlisting}
SELECT customer_id, (SELECT SUM(amount) FROM sales s WHERE s.customer_id = sales.customer_id) as total_sales FROM sales GROUP BY customer_id;
    \end{lstlisting}

  \item[Query: Find the average salary for each department.] ~\\
    \textbf{SQL1:}
    \begin{lstlisting}
SELECT dept_id, AVG(salary) as avg_salary FROM employees GROUP BY dept_id;
    \end{lstlisting}
    \textbf{SQL2:}
    \begin{lstlisting}
SELECT dept_id, (SELECT AVG(salary) FROM employees e WHERE e.dept_id = employees.dept_id) as avg_salary FROM employees GROUP BY dept_id;
    \end{lstlisting}
\end{description}

\subsubsection{Filtering Methods}
Using different filtering methods to achieve the same result.

\begin{description}
  \item[Query: Retrieve all products that are either in category 'Electronics' or cost more than \$100.] ~\\
    \textbf{SQL1:}
    \begin{lstlisting}
SELECT * FROM products WHERE category = 'Electronics' OR price > 100;
    \end{lstlisting}
    \textbf{SQL2:}
    \begin{lstlisting}
SELECT * FROM products WHERE category = 'Electronics'
UNION
SELECT * FROM products WHERE price > 100;
    \end{lstlisting}

  \item[Query: Get all employees who work in 'HR' or 'Finance'.] ~\\
    \textbf{SQL1:}
    \begin{lstlisting}
SELECT * FROM employees WHERE dept = 'HR' OR dept = 'Finance';
    \end{lstlisting}
    \textbf{SQL2:}
    \begin{lstlisting}
SELECT * FROM employees WHERE dept = 'HR'
UNION
SELECT * FROM employees WHERE dept = 'Finance';
    \end{lstlisting}
\end{description}

\subsubsection{CASE vs. Multiple WHERE Clauses}
Using CASE statements versus multiple WHERE clauses (typically with UNION) to achieve the same result.

\begin{description}
  \item[Query: Get the employee names and their status (Active/Inactive).] ~\\
    \textbf{SQL1:}
    \begin{lstlisting}
SELECT name, CASE WHEN active = 1 THEN 'Active' ELSE 'Inactive' END as status FROM employees;
    \end{lstlisting}
    \textbf{SQL2:}
    \begin{lstlisting}
SELECT name, 'Active' as status FROM employees WHERE active = 1
UNION ALL
SELECT name, 'Inactive' as status FROM employees WHERE active = 0;
    \end{lstlisting}

  \item[Query: Find the products with their availability status (In Stock/Out of Stock).] ~\\
    \textbf{SQL1:}
    \begin{lstlisting}
SELECT product_name, CASE WHEN stock > 0 THEN 'In Stock' ELSE 'Out of Stock' END as availability FROM products;
    \end{lstlisting}
    \textbf{SQL2:}
    \begin{lstlisting}
SELECT product_name, 'In Stock' as availability FROM products WHERE stock > 0
UNION ALL
SELECT product_name, 'Out of Stock' as availability FROM products WHERE stock = 0;
    \end{lstlisting}
\end{description}

\subsubsection{ORDER BY Clauses}
Using different ORDER BY clauses to achieve the same result. (This can be tricky and highly dependent on data types and specific SQL dialect features).

\begin{description}
  \item[Query: List all products ordered by price from highest to lowest.] ~\\
    \textbf{SQL1:}
    \begin{lstlisting}
SELECT * FROM products ORDER BY price DESC;
    \end{lstlisting}
    \textbf{SQL2 (Illustrative, may not be universally equivalent or good practice):}
    \begin{lstlisting}
SELECT * FROM products ORDER BY price * -1 ASC;
    \end{lstlisting}

  \item[Query: Get all employees ordered by their hire date from most recent to oldest.] ~\\
    \textbf{SQL1:}
    \begin{lstlisting}
SELECT * FROM employees ORDER BY hire_date DESC;
    \end{lstlisting}
    \textbf{SQL2 (Illustrative, highly dialect-specific if negative sign works on dates):}
    \begin{lstlisting}
-- SELECT * FROM employees ORDER BY -hire_date;
-- This syntax is not standard SQL for dates and might not work or might have unintended behavior.
-- A more robust equivalent would be identical to SQL1 or rely on specific date functions if available.
-- For illustration, if hire_date was a numeric representation (e.g., epoch seconds), then ORDER BY -hire_date could work.
    \end{lstlisting}
\end{description}

\subsubsection{EXISTS vs. JOIN}
Using EXISTS versus JOIN clauses to achieve the same result.

\begin{description}
  \item[Query: Find customers who have placed at least one order.] ~\\
    \textbf{SQL1:}
    \begin{lstlisting}
SELECT customer_name FROM customers WHERE EXISTS (SELECT 1 FROM orders WHERE customers.customer_id = orders.customer_id);
    \end{lstlisting}
    \textbf{SQL2:}
    \begin{lstlisting}
SELECT DISTINCT c.customer_name
FROM customers c
JOIN orders o ON c.customer_id = o.customer_id;
    \end{lstlisting}

  \item[Query: Get the names of students who are enrolled in any course.] ~\\
    \textbf{SQL1:}
    \begin{lstlisting}
SELECT student_name FROM students WHERE EXISTS (SELECT 1 FROM enrollments WHERE students.student_id = enrollments.student_id);
    \end{lstlisting}
    \textbf{SQL2:}
    \begin{lstlisting}
SELECT DISTINCT s.student_name
FROM students s
JOIN enrollments e ON s.student_id = e.student_id;
    \end{lstlisting}
\end{description}

\subsection{Patterns of Semantically Inequivalent SQL Queries}

\subsubsection{Incorrect JOIN Conditions}
Using incorrect or incomplete JOIN conditions that lead to incorrect results.
\begin{description}
  \item[Query: Retrieve all orders with their corresponding customer names.] ~\\
    \textbf{SQL1 (Correct):}
    \begin{lstlisting}
SELECT orders.order_id, customers.customer_name
FROM orders
INNER JOIN customers ON orders.customer_id = customers.customer_id;
    \end{lstlisting}
    \textbf{SQL2 (Incorrect):}
    \begin{lstlisting}
SELECT orders.order_id, customers.customer_name
FROM orders
INNER JOIN customers ON orders.order_id = customers.customer_id; -- Incorrect join key
    \end{lstlisting}
\end{description}

\subsubsection{Incorrect WHERE Clauses}
Using incorrect or incomplete WHERE clauses that filter results incorrectly.
\begin{description}
  \item[Query: Find all products that cost more than \$100.] ~\\
    \textbf{SQL1 (Correct):}
    \begin{lstlisting}
SELECT * FROM products WHERE price > 100;
    \end{lstlisting}
    \textbf{SQL2 (Incorrect):}
    \begin{lstlisting}
SELECT * FROM products WHERE price >= 100; -- Incorrect operator
    \end{lstlisting}
\end{description}

\subsubsection{Incorrect Aggregation}
Using incorrect aggregation functions or grouping that lead to incorrect results.
\begin{description}
  \item[Query: Get the total sales amount for each customer.] ~\\
    \textbf{SQL1 (Correct):}
    \begin{lstlisting}
SELECT customer_id, SUM(amount) as total_sales
FROM sales
GROUP BY customer_id;
    \end{lstlisting}
    \textbf{SQL2 (Incorrect):}
    \begin{lstlisting}
SELECT customer_id, AVG(amount) as total_sales -- Incorrect aggregate function
FROM sales
GROUP BY customer_id;
    \end{lstlisting}
\end{description}

\subsubsection{Misuse of DISTINCT and GROUP BY}
Incorrect use of DISTINCT and GROUP BY clauses leading to incorrect results.
\begin{description}
  \item[Query: Get the unique list of product categories.] ~\\
    \textbf{SQL1 (Correct):}
    \begin{lstlisting}
SELECT DISTINCT category FROM products;
    \end{lstlisting}
    \textbf{SQL2 (Incorrect):}
    \begin{lstlisting}
SELECT category FROM products GROUP BY category, price; -- Incorrect grouping, changes semantics
    \end{lstlisting}
\end{description}

\subsubsection{Incorrect Subqueries}
Using subqueries incorrectly that lead to incorrect results.
\begin{description}
  \item[Query: List the names of customers who have placed an order.] ~\\
    \textbf{SQL1 (Correct):}
    \begin{lstlisting}
SELECT customer_name
FROM customers
WHERE customer_id IN (SELECT customer_id FROM orders);
    \end{lstlisting}
    \textbf{SQL2 (Incorrect - assuming subquery might return multiple rows for '='):}
    \begin{lstlisting}
SELECT customer_name
FROM customers
WHERE customer_id = (SELECT customer_id FROM orders); -- Incorrect if subquery returns >1 row
    \end{lstlisting}
\end{description}

\subsubsection{Incorrect Filtering with AND/OR}
Misusing AND/OR in WHERE clauses that lead to incorrect results.
\begin{description}
  \item[Query: Retrieve all products that are in category 'Electronics' AND cost more than \$100.] ~\\
    \textbf{SQL1 (Correct):}
    \begin{lstlisting}
SELECT * FROM products WHERE category = 'Electronics' AND price > 100;
    \end{lstlisting}
    \textbf{SQL2 (Incorrect):}
    \begin{lstlisting}
SELECT * FROM products WHERE category = 'Electronics' OR price > 100; -- Incorrect logical operator
    \end{lstlisting}
\end{description}

\subsubsection{Incorrect ORDER BY Clauses}
Using incorrect ORDER BY clauses that lead to incorrect sorting.
\begin{description}
  \item[Query: List all products ordered by price from highest to lowest.] ~\\
    \textbf{SQL1:}
    \begin{lstlisting}
SELECT * FROM products ORDER BY price DESC;
    \end{lstlisting}
    \textbf{SQL2 (Incorrect):}
    \begin{lstlisting}
SELECT * FROM products ORDER BY price ASC; -- Incorrect sort order
    \end{lstlisting}
\end{description}

\subsubsection{Misuse of Functions}
Using incorrect functions or misusing functions that lead to incorrect results.
\begin{description}
  \item[Query: Find the length of each product name. (Assuming LENGTH and CHAR\_LENGTH might differ for some charsets\/DBs)] ~\\
    \textbf{SQL1 (Correct for byte length, or char length if DB treats them same):}
    \begin{lstlisting}
SELECT product_name, LENGTH(product_name) as name_length
FROM products;
    \end{lstlisting}
    \textbf{SQL2 (Potentially Incorrect if expecting byte length and CHA\_LENGTH gives char count for multi-byte chars):}
    \begin{lstlisting}
SELECT product\_name, CHAR\_LENGTH(product_name) as name_length
FROM products;
    \end{lstlisting}
\end{description}

\section{Prompt Examples}
\label{apx:prompt_examples}

This section provides examples of prompts used to guide the LLM in assessing SQL equivalence.

\subsection{Basic Prompt example for Equivalence Assessment}

\begin{nicebox}{Basic Prompt Template}
You are a database analyst and  an SQL expert. Your task is to determine if two given SQL queries are semantically equivalent. The queries are intended for [Application name].

Please follow the following rules:

[rules]

and below are some examples:

[examples]
\

NL query:
[NL query]

Query 1:
[SQL1]

Query 2:
[SQL2]

Please think step by step, and provide your reasoning before giving the final answer. Output your answer in the following json format:
\begin{lstlisting}
{
  "reasoning": "Your reasoning here",
  "equivalence": "equivalent" or "not equivalent"
}
\end{lstlisting}

\end{nicebox}

\subsection{Improved Prompt example for Equivalence Assessment}

\begin{nicebox}{Improved Prompt Template}
Task: Determine if two given SQL queries are semantically equivalent. The queries are intended
for [Application name]. 

The criteria of equivalence of two SQL queries are defined as follows:
[detailed criteria of equivalence]

and the criteria of the assessment are:
[detailed criteria of grading]

Please follow the following rules:

[rules]

and below are some examples:

[examples with Chain-of-Thought]
\
NL query:
[NL query]

Query 1:
[SQL1]

Query 2:
[SQL2]

[Optional Schema Information Here]

[Optional Query Execution Results Here]

Please think step by step, and provide your reasoning before giving the final answer. Output your answer in the following json format:
\begin{lstlisting}
{
  "reasoning": "Your reasoning here",
  "overall accessment": "equivalent"/"minor difference"/"significant difference"/"not equivalent"/"undermined"
}
\end{lstlisting}

\end{nicebox}

\subsection{Prompt for Miniature \& Mull Strategy}
\label{apx:prompt_miniature_mull}
\begin{nicebox}{Miniature \& Mull Prompt Template \cite{zhao2024llmsqlsolverllmsdeterminesql}}
/* Given the following two SQL queries Q1 and Q2 */

SQL1:[SQL1]
SQL2:[SQL2]

/* And the following database schema: */

[schema]

/* Are Q1 and Q2 semantically equivalent?

1. Try one example database and check the output table of Q1 and Q2. Database is case-sensitive when comparing string values.

2. If the outputs are identical, adjust the database to see how output tables of Q1 and Q2 change.

3. After evaluating whether there exists a database such Q1 and Q2 output different tables, write your answer using format decision = "equivalent" or decision = "inequivalent".
*/

/* Let's think step by step. */

1. Consider the following example database instance, which is string value case-sensitive, and execute Q1 and Q2.
\end{nicebox}

\subsection{Combined Prompt example: Detailed Assessment with Miniature \& Mull Reasoning}
\label{apx:prompt_combined_m_and_m}
\begin{nicebox}{Combined Prompt: Detailed Assessment with Miniature \& Mull}
Task: Determine if two given SQL queries are semantically equivalent. The queries are intended
for [Application name].

The criteria of equivalence of two SQL queries are defined as follows:
[detailed criteria of equivalence]

And the criteria of the assessment are:
[detailed criteria of grading]

Please follow the following rules:
[rules]

And below are some examples:
[examples with Chain-of-Thought, potentially showing M\&M style reasoning]

NL query:
[NL query]

Query 1 (Q1):
[SQL1]

Query 2 (Q2):
[SQL2]

Database Schema:
[schema]

[Optional: Other Context or Query Execution Results if not using M\&M exclusively for execution simulation]

To determine semantic equivalence, please adopt the following "Miniature \& Mull" thinking process:
1. Try one example database instance based on the provided schema. Execute Q1 and Q2 on this database. Remember the database is case-sensitive when comparing string values.
2. If the outputs of Q1 and Q2 are identical on this first instance, adjust the database (e.g., add/remove/modify rows or values) to create a new instance. Re-execute Q1 and Q2 and observe how their output tables change.
3. Repeat step 2 if necessary, trying to find a database instance where Q1 and Q2 produce different output tables.

After thoroughly applying this thinking process, please provide your reasoning and final assessment.
Output your answer in the following JSON format:
\begin{lstlisting}
{
  "reasoning": "Your step-by-step reasoning, including the database instances considered and the outputs of Q1 and Q2 on them, and how this led to your conclusion.",
  "overall_assessment": "equivalent" / "minor_difference" / "significant_difference" / "not_equivalent" / "undetermined"
}
\end{lstlisting}
\end{nicebox}
\section{Evaluation Algorithms}
\label{apx:algorithms}
\begin{algorithm}[tb]
   \caption{SQL Equivalence Evaluation Pipeline (Basic)}
   \label{alg:sql_eval_pipeline}
\begin{algorithmic}[1] 
   \STATE {\bfseries Input:} SQL Query Pair $(Q_1, Q_2)$, Database Schema $S$ (optional), Execution Results $R$ (optional)
   \STATE {\bfseries Output:} Equivalence Judgment (Equivalent, Not Equivalent, Unstable)

   \STATE \COMMENT{Step 1: Preprocessing}
   \STATE $Q_1' \leftarrow \text{Preprocess}(Q_1)$ \COMMENT{Standardize formatting, e.g., date/time, filters}
   \STATE $Q_2' \leftarrow \text{Preprocess}(Q_2)$

   \STATE \COMMENT{Step 2: String-based Matching}
   \IF{$\text{ExactMatch}(Q_1', Q_2')$}
     \STATE \textbf{return} Equivalent
   \ENDIF
   \IF{$\text{ExactSetMatch}(Q_1', Q_2', S)$}
     \STATE \textbf{return} Equivalent
   \ENDIF
   \COMMENT{Note: If string-based methods determine clear inequivalence, could return Not Equivalent here, or proceed to LLM for deeper analysis/explanation.}

   \STATE \COMMENT{Step 3: LLM-based Evaluation (Standard Prompting)}
   \STATE Initialize $judgments \leftarrow []$
   \STATE Set $default\_runs \leftarrow 3$, $max\_runs \leftarrow 5$
   \STATE Set $current\_runs \leftarrow default\_runs$

   \FOR{$i=1$ {\bfseries to} $current\_runs$}
     \STATE $prompt \leftarrow \text{ConstructPrompt}(Q_1', Q_2', S, R)$ \COMMENT{See Appendix~\ref{apx:prompt_examples} for basic prompt}
     \STATE $judgment_i \leftarrow \text{LLM\_Query}(prompt, \text{GPT-4 model})$
     \STATE Add $judgment_i$ to $judgments$
   \ENDFOR

   \IF{all $judgment \in judgments$ are identical}
     \STATE \textbf{return} $judgments[0]$
   \ELSE
     \STATE \COMMENT{Judgments vary, result is unstable}
     \IF{$current\_runs < max\_runs$}
        \STATE \COMMENT{Optional: Increase runs for unstable cases}
        \FOR{$i=current\_runs+1$ {\bfseries to} $max\_runs$}
            \STATE $prompt \leftarrow \text{ConstructPrompt}(Q_1', Q_2', S, R)$
            \STATE $judgment_i \leftarrow \text{LLM\_Query}(prompt, \text{GPT-4 model})$
            \STATE Add $judgment_i$ to $judgments$
        \ENDFOR
        \STATE $final\_judgment \leftarrow \text{MajorityVote}(judgments)$
        \IF{$final\_judgment$ is conclusive}
            \STATE \textbf{return} $final\_judgment$
        \ELSE
            \STATE \textbf{return} Unstable \COMMENT{Even after more runs}
        \ENDIF
     \ELSE
        \STATE \textbf{return} Unstable
     \ENDIF
   \ENDIF
\end{algorithmic}
\end{algorithm}

\begin{algorithm}[tb]
   \caption{Improved SQL Equivalence Evaluation Pipeline with Query Rewrite and Miniature \& Mull Strategy}
   \label{alg:improved_sql_eval_pipeline}
\begin{algorithmic}[1] 
   \STATE {\bfseries Input:} SQL Query Pair $(Q_1, Q_2)$, Database Schema $S$ (optional)
   \STATE {\bfseries Output:} Equivalence Judgment (Equivalent, Not Equivalent, Unstable)

   \STATE \COMMENT{Step 1: Preprocessing}
   \STATE $Q_{1p} \leftarrow \text{Preprocess}(Q_1)$ \COMMENT{Standardize formatting}
   \STATE $Q_{2p} \leftarrow \text{Preprocess}(Q_2)$

   \STATE \COMMENT{Step 2: Query Rewrite for Subqueries}
   \STATE $Q_{1r} \leftarrow \text{RewriteSubqueriesToJoins}(Q_{1p})$ \COMMENT{e.g., to LEFT JOIN}
   \STATE $Q_{2r} \leftarrow \text{RewriteSubqueriesToJoins}(Q_{2p})$

   \STATE \COMMENT{Step 3: String-based Matching (on potentially rewritten queries)}
   \IF{$\text{ExactMatch}(Q_{1r}, Q_{2r})$}
     \STATE \textbf{return} Equivalent
   \ENDIF
   \IF{$\text{ExactSetMatch}(Q_{1r}, Q_{2r}, S)$}
     \STATE \textbf{return} Equivalent
   \ENDIF

   \STATE \COMMENT{Step 4: LLM-based Evaluation with Miniature \& Mull}
   \STATE Initialize $judgments \leftarrow []$
   \STATE Set $default\_runs \leftarrow 3$, $max\_runs \leftarrow 5$
   \STATE Set $current\_runs \leftarrow default\_runs$

   \FOR{$i=1$ {\bfseries to} $current\_runs$}
     \STATE $prompt \leftarrow \text{ConstructMiniatureAndMullPrompt}(Q_{1r}, Q_{2r}, S)$ \COMMENT{Instructs LLM to create DB, execute, modify DB, re-execute, compare}
     \STATE $judgment_i \leftarrow \text{LLM\_Query}(prompt, \text{GPT-4 model})$
     \STATE Add $judgment_i$ to $judgments$
   \ENDFOR

   \IF{all $judgment \in judgments$ are identical}
     \STATE \textbf{return} $judgments[0]$
   \ELSE
     \STATE \COMMENT{Judgments vary, result is unstable}
     \IF{$current\_runs < max\_runs$}
        \STATE \COMMENT{Optional: Increase runs for unstable cases}
        \FOR{$i=current\_runs+1$ {\bfseries to} $max\_runs$}
            \STATE $prompt \leftarrow \text{ConstructMiniatureAndMullPrompt}(Q_{1r}, Q_{2r}, S)$
            \STATE $judgment_i \leftarrow \text{LLM\_Query}(prompt, \text{GPT-4 model})$
            \STATE Add $judgment_i$ to $judgments$
        \ENDFOR
        \STATE $final\_judgment \leftarrow \text{MajorityVote}(judgments)$
        \IF{$final\_judgment$ is conclusive}
            \STATE \textbf{return} $final\_judgment$
        \ELSE
            \STATE \textbf{return} Unstable \COMMENT{Even after more runs}
        \ENDIF
     \ELSE
        \STATE \textbf{return} Unstable
     \ENDIF
   \ENDIF
\end{algorithmic}
\end{algorithm}

\end{document}